\documentclass[letterpaper, 10 pt, conference]{ieeeconf}
\IEEEoverridecommandlockouts
\usepackage{cite}
\usepackage{amsmath}
\usepackage{amsfonts}
\usepackage{amssymb}
\usepackage{nicematrix}
\usepackage{url}
\usepackage{booktabs}
\usepackage{caption}
\usepackage{xcolor}
\usepackage{multirow}
\usepackage{makecell}
\usepackage{xcolor}
\usepackage[hidelinks]{hyperref}
\usepackage{booktabs,tabularx,array,makecell}
\newcolumntype{L}[1]{>{\raggedright\arraybackslash}p{#1}}
\newcolumntype{C}[1]{>{\centering\arraybackslash}p{#1}}

\usepackage[normalem]{ulem}

\newcommand{\state}{\mathbf{x}}
\newcommand{\statedot}{\mathbf{\dot{x}}}
\newcommand{\p}{\mathbf{p}}
\newcommand{\linvel}{\dot{\mathbf{p}}}
\newcommand{\ori}{\boldsymbol{\Theta}}
\newcommand{\angvel}{\boldsymbol{\omega}}

\newcommand{\inp}{\mathbf{u}}
\newcommand{\force}{\mathbf{F}}
\newcommand{\moment}{\mathbf{M}}
\newcommand{\rot}{\mathbf{R}_b}
\newcommand{\pfoot}{\mathbf{p}_\text{foot}}

\newcommand{\vdes}{\dot{\mathbf{p}}^{\text{ref}}}
\newcommand{\pfdes}{\mathbf{p}^\text{ref}_\text{foot}}
\newcommand{\foothold}{\mathbf{p}_\text{foothold}}
\newcommand{\zdes}{\mathrm{p}^{\text{ref}, \mathrm{z}}}
\newcommand{\phip}{\mathbf{p}_\text{hip}}
\newcommand{\moI}{I_\mathbf{w}}
\newcommand{\g}{\mathbf{g}}

\newcommand{\dynamics}{\mathbf{f}(\mathbf{x}, \mathbf{u})}
\newcommand{\nominaldynamics}{\widetilde{\mathbf{f}}(\mathbf{x}, \mathbf{u})}
\newcommand{\statespace}{\statedot = \mathbf{A}\state+\mathbf{B}\inp}
\newcommand{\Bezier}{\mathcal{B}(\p_0, \p_1, \p_2, \mathbf{p}_\text{foothold}, \phi_t)}

\title{\LARGE \bf
RL-augmented Adaptive Model Predictive Control for Bipedal Locomotion over Challenging Terrain
}

\author{Junnosuke Kamohara$^{1}$, 
        Feiyang Wu$^{1}$, 
        Chinmayee Wamorkar$^{1}$, 
        Seth Hutchinson$^{2}$, 
        Ye Zhao$^{1}$%
\thanks{
    $^{1}$ Junnosuke Kamohara, Feiyang Wu, Chinmayee Wamorkar, and Ye Zhao are with Georgia Institute of Technology, GA 30332, USA. {\tt\small \{jkamohara3, feiyangwu, cwamorkar3, yezhao\}@gatech.edu} \newline \indent
    $^{2}$ Seth Huchinson is with Northeastern University, 360 Huntington Ave, Boston, MA 02115, USA. {\tt\small s.hutchinson@northeastern.edu}}%
}

\begin{document}

\maketitle
\thispagestyle{empty}
\pagestyle{empty}

\begin{abstract}
Model predictive control (MPC) has demonstrated effectiveness for humanoid bipedal locomotion; however, its applicability in challenging environments, such as rough and slippery terrain, is limited by the difficulty of modeling terrain interactions. 
In contrast, reinforcement learning (RL) has achieved notable success in training robust locomotion policies over diverse terrain, yet it lacks guarantees of constraint satisfaction and often requires substantial reward shaping. 
Recent efforts in combining MPC and RL have shown promise of taking the best of both worlds, but they are primarily restricted to flat terrain or quadrupedal robots. 

\noindent In this work, we propose an RL-augmented MPC framework tailored for bipedal locomotion over rough and slippery terrain. 
Our method parametrizes three key components of single-rigid-body-dynamics-based MPC: system dynamics, swing leg controller, and gait frequency.
We validate our approach through bipedal robot simulations in NVIDIA IsaacLab across various terrains, including stairs, stepping stones, and low-friction surfaces.
Experimental results demonstrate that our RL-augmented MPC framework produces significantly more adaptive and robust behaviors compared to baseline MPC and RL.
Project page: \url{https://rl-augmented-mpc.github.io/rlaugmentedmpc/}
\end{abstract}

\section{Introduction}
Legged locomotion conventionally employs model-based controllers (MBCs), particularly Model Predictive Control (MPC), due to their optimization-based constraint satisfaction~\cite{kohler2019linear, feller2016relaxed}. 
While whole-body dynamics models~\cite{khazoom2024tailoring} are more accurate, 
researchers use simplified models~\cite{kajita20013d, dai2014whole, di2018dynamic, li2021force} for computational efficiency and consequently suffer from model mismatch due to simplification of the dynamics. 
As a result, simplified models exhibit poorer tracking accuracy and instability, particularly during contact~\cite{kim2025modular}. 
Additionally, MPC with simplified dynamics usually requires predefined contact sequence and swing leg trajectory, which limits its adaptivity to diverse terrains.
Overall, the deterministic but inaccurate dynamic model and manual constraint design of MPC restrict its robustness and versatility, limiting its applicability to diverse terrains in the real world. 

In contrast, learning-based controls (LBC), exemplified by Reinforcement Learning (RL) methods, have gained wide attention for their robustness and agility~\cite{lee2020learning,rudin2022learning,miki2022learning,wu2025learn,jung2025ppf}.
By training policies parameterized by neural networks, RL policies can achieve zero-shot transfer from simulation to reality.
However, training robust policies requires substantial environmental interactions and extensive reward shaping. 
Furthermore, RL policies lack explicit constraint satisfaction because of the absence of explicit constraints.

Motivated by the unique advantages of both sides, recent years have witnessed a surge of methods combining model-based and learning-based approaches, leveraging the safety offered by MPC's explicit constraints as well as powerful reactive behaviors offered by RL~\cite{reiter2025synthesis, gu2025humanoid}.
In legged robotics, there are two main threads of combination. 
The first thread uses MPC within a policy. 
Recent works either adopt a \textit{hierarchical architecture}, where RL parametrizes MPC's components, including system dynamics, center of mass reference trajectory, and gait frequency~\cite{pandala2022robust, kim2025modular, xie2022glide, yang2023cajun}; or follows a \textit{parallel architecture}, where RL policies refine MPC outputs by adding corrective actions such as footholds and joint commands~\cite{bang2024rl, chen2024learning-usc, cheng2025rambo}. 
Another thread uses MPC as an expert policy, training the policy through behavior cloning or RL with imitation loss to increase sample efficiency and motion accuracy~\cite{lee2024integrating, jung2025ppf, liu2024opt2skill, jenelten2024dtc}.
Each of these designs carries trade-offs: 
MPC as an expert improves training efficiency by imitating MPC motions, yet it incurs significant computational overhead during training due to repeated optimization solves, making training in parallelized RL environments particularly challenging~\cite{jenelten2024dtc}. 
While parallel architectures offer flexibility by directly augmenting MPC outputs, they raise safety concerns since the RL policy bypasses feasibility constraints from optimization. 
Hierarchical architectures, in contrast, preserve the optimization structure and computational complexity, as the policy is evaluated before solving the optimization problem. 
This ensures the feasibility and constraint satisfaction within the optimization framework.

Despite these advances, most combined approaches for bipedal locomotion remain limited to flat terrain, as prior works primarily emphasize improving tracking accuracy rather than adaptability~\cite{bang2024rl,cheng2025rambo}, leaving integration of MPC and RL for rough-terrain-adaptive locomotion unexplored.
In this work, we aim to enhance the adaptability of humanoid locomotion, enabling responsive and robust behaviors in the face of terrain disturbances.
We leverage a hierarchical method that augments MPC via RL by incorporating rich whole-body information into the simplified system model, adjusting the gait frequency to modulate step length, and modifying the swing foot trajectory to improve robustness against challenging terrain. 
We focus on addressing the limitations of MPC with simplified dynamics: model mismatch, predefined swing leg curve, and static gait frequency. 
The RL policy learns residual dynamics through whole-body dynamics simulation, as well as swing leg curve parameters, including apex height and control points, and dynamic gait frequency within one locomotion cycle.

These learned adaptations enable reactive behaviors, including recovery from foot entrapment and severe slippage. 
We implement our method on bipedal locomotion tasks with the HECTOR robot~\cite{li2023dynamic} in NVIDIA IsaacLab, a state-of-the-art GPU-accelerated simulator~\cite{mittal2023orbit}.
Our framework significantly improves robustness against disturbances on diverse terrains, including slippery surfaces, stairs, and stepping stones. 
Additionally, we conduct ablation studies on the three residual modules to analyze the contribution of each component.

\begin{figure}[t]
    \noindent\begin{minipage}{\linewidth}
        \centering
        \includegraphics[width=1.0\linewidth]{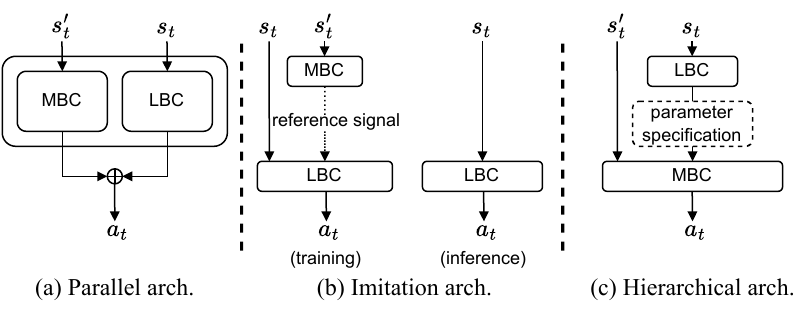}
        \vspace{-5mm}
    \end{minipage}
    \caption{Overview of existing MPC and RL combined approaches. 
        MBC and LBC take state $s^{\prime}_t$ and observation $s_t$ as inputs. 
        The final control output is $a_t$. }
    \label{fig:fig1}
    \vspace{-5mm}
\end{figure}

\begin{figure*}[t]
  \centering
  \includegraphics[width=17.5cm]{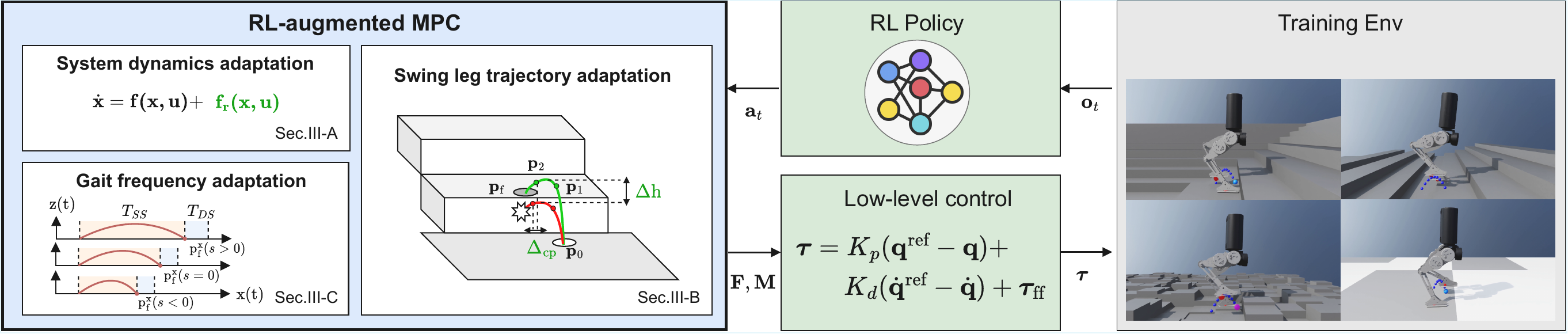}
  \caption{
    System architecture of our proposed method. 
    The RL policy augments three key modules in the single rigid body MPC. 
    The low level controller then tracks the contact force $\force$ and moment $\moment$ computed by MPC to generate joint torque commands. 
    We leverage the whole-body dynamics of the robot to train the policy across diverse terrains, as illustrated on the right side of the figure. 
    }
  \label{fig:fig2}
  \vspace{-5mm}
\end{figure*}

\section{Related work}



Recent works have explored how to combine MPC and RL controllers~\cite{reiter2025synthesis, gu2025humanoid}. 
Existing literature in legged locomotion follows three types of architectures: parallel, imitation, and hierarchical architecture, as shown in Fig.~\ref{fig:fig1}.

\noindent \textbf{Parallel architecture: }
With a parallel architecture, MPC's output is refined by a learning-based module to improve the performance.
Learning-based controllers are usually trained to output the residual action, which is then added to the MPC output to form the final control. 
A2C-MPC~\cite{gupta2024reinforcement} outputs forward acceleration and angular velocity in the steering angle to improve tracking in off-road navigation problems. 
Authors in~\cite{chen2024learning-usc,cheng2025rambo} augment joint commands in an attempt to achieve agile locomotion on rough terrain. 
In bipedal locomotion, authors in~\cite{bang2024rl} refine suboptimal footholds from MPC by training a residual foot placement planner using RL.
While these approaches are intuitive and can directly override control inputs, they may fail to produce stable walking motions due to the lack of explicit constraint satisfaction. 

\noindent\textbf{Imitation architecture: }
This approach employs MPC to guide the RL policy during training~\cite{jung2025ppf, liu2024opt2skill, jenelten2024dtc}, leveraging trajectory optimization solutions as additional supervision signals. 
Model-based reference generation can help accelerate training, but it also has significant drawbacks.
Training with offline-generated trajectory limits the policy's generalizability, as the trained policy only imitates the generated trajectories~\cite{jung2025ppf, liu2024opt2skill}. 
Employing MPC as an online trajectory generator can yield a more generalizable policy, but the MPC in this setup is often overly simplified or restricted due to heavy computational overhead~\cite{jenelten2024dtc}.

\noindent \textbf{Hierarchical architecture: }
With a hierarchical architecture, RL specifies MPC's parameters, and MPC produces the final control inputs. 
This line of research often leverages residual estimation~\cite{shi2019neural, pandala2022robust, kim2025modular} and learning-based planners~\cite{xie2022glide, yang2023cajun}. 
For residual estimation, the authors in~\cite{pandala2022robust, chen2024learning-usc} estimate the uncertainty sets of system dynamics via RL, achieving locomotion on diverse terrain. 
Another approach~\cite{shi2019neural, kim2025modular} employs supervised learning to estimate the residual dynamics from offline proprioceptive data. 
For learning-based planners, CAjun~\cite{yang2023cajun} trains an RL policy that outputs Center of Mass (CoM) velocity commands, gait frequency, and refined footholds to achieve jumping. 
GLiDE~\cite{xie2022glide} estimates reference CoM acceleration, which is tracked by a quadratic programming-based force controller. 
Nonetheless, for bipedal robots, hierarchical training with RL and MPC remains an area of investigation. 
Inspired by this, our proposed RL-augmented MPC for humanoid robots augments the system dynamics through RL training, while simultaneously enabling adaptive swing leg trajectory generation and gait frequency modulation to further enhance versatility and robustness.

\noindent\textbf{Bipedal locomotion over challenging terrain: }
Recent works have demonstrated successful bipedal locomotion over diverse rough terrains~\cite{liao2025berkeley, radosavovic2024real, liu2024opt2skill,jung2025ppf, wu2025learn}. 
However, RL-based policies raise safety concerns and demand significant reward shaping efforts. 
MPC and RL combined approaches aim to address these issues, but they are largely limited to quadrupedal locomotion, and the evaluated terrains are restricted to a short staircase~\cite{chen2024learning-usc}, stepping stones~\cite{xie2022glide}, and gravel and grass~\cite{pandala2022robust}. 
Still, it remains to be seen if these works can succeed on more general rough terrains like stairs with a larger number of steps or terrains with random elevation patterns. 

\section{Preliminaries}
\newcommand{\R}{\mathbb{R}}
\newcommand{\relcontactpos}{\mathbf{r}}

The following section briefly introduces the single rigid body dynamics (SRBD) MPC controller~\cite{di2018dynamic, li2021force}. 

\subsection{Single rigid body dynamics MPC}
The SRBD can be formulated as follows:
\begin{align}
    \begin{array}{l}
    \statedot = \dynamics =
    \begin{bmatrix}
        \linvel
        \\
        \rot\angvel
        \\
        \sum_i \force_i/m + \g
        \\
        \moI^{-1} \sum_i (\relcontactpos_i \times \force_i + \moment_i),
    \end{bmatrix}
    \end{array}
    \label{eq:srbd-dynamics}
\end{align}
where $\state = [\p, \ori, \linvel, \angvel] \in \R^{12}$ is the state, and $\inp = [\force_1, \force_2, \moment_1, \moment_2] \in \R^{12}$ is the control input. 
The state $\mathbf{x}$ consists of the center of mass (CoM) position $\mathbf{p} \in \R^3$, orientation $\ori \in \R^3$, linear velocity $\linvel \in R^3$, and angular velocity $\angvel \in \R^3$. 
The control input $\inp$ represents the ground reaction force $\force \in \R^3$ and the reaction wrench $\moment \in \R^3$, where $i \in \{0, 1\}$ is the index of leg in contact.
Additionally, $\g \in \R^3$ is the gravity vector, $m$ is the lumped robot mass, and $\moI \in \R^{3\times3}$ is the robot's centroidal inertia w.r.t the global coordinate, $\relcontactpos_i \in \R^3$ is the vector from the CoM to the $i^{\rm th}$ contact point, and $\rot \in \R^{3\times3}$ is the world-to-base coordinate transformation matrix. 
The nonlinear terms $\rot \angvel$ and $\relcontactpos_i\times \force_i$ are linearized at the current state to obtain linear dynamics.
We refer to this linearized state dynamics as the nominal dynamics $\nominaldynamics\!:\statespace$, with an augmented state $\state = [\p,\ori,\linvel,\angvel,1] \in \R^{13}$ and state-space matrices $\mathbf{A} \in \R^{13\times13}$ and $\mathbf{B} \in \R^{13\times12}$. 
We include the gravity term as an additional state variable to cast the dynamics into state-space form~\cite{di2018dynamic, li2021force}, and we continue to denote the augmented state as $\state$, by slight abuse of notation.

\newcommand{\decisionset}{\boldsymbol{\mathcal{X}}}
\newcommand{\Qweight}{\mathbf{Q}_k}
\newcommand{\Rweight}{\mathbf{R}_k}
\newcommand{\Amat}{\mathbf{A}_k}
\newcommand{\Bmat}{\mathbf{B}_k}
\newcommand{\SelMat}{\mathbf{S}}
\newcommand{\FootRotMat}{\mathbf{R}_{\text{f}, i}}
\newcommand{\contactstate}{\mathrm{c}_{k,i}}

Using the linearized dynamics, the convex MPC controller~\cite{li2021force} solves a
constrained optimization problem that computes the contact wrench to track the reference trajectory:
\begin{subequations}\label{eq:mpc}
    \begin{align}
    \min_{\decisionset} \; 
    &\textstyle{\sum}_{k=0}^{N-1}(\state_{k+1}-\state_{k+1}^{\text{ref}})^\top \Qweight (\state_{k+1}-\state_{k+1}^{\text{ref}})
    + \inp_k^\top \Rweight \inp_k \nonumber
    \\
    \text{s.t. } \  
    &\state_{k+1} = \Amat \state_k + \Bmat \inp_k\label{eq:mpc-b} \\
     &(1-\contactstate)\,\inp_{k,i} = \mathbf{0}\label{eq:mpc-c} \\
     &|\mathrm{F^x}_{k,i}|, |\mathrm{F^y}_{k,i}| \leq \mu |\mathrm{F^z}_{k,i}| \label{eq:mpc-d} \\
     & 0 \leq \mathrm{F^z}_{k,i} \leq \mathrm{F}_{\max} \label{eq:mpc-e} \\
    &-l_t\SelMat_\mathrm{z}\FootRotMat^\top\force_{k,i} \leq \SelMat_\mathrm{y}\FootRotMat^\top\moment_{k,i} \leq l_h\SelMat_\mathrm{z}\FootRotMat^\top\force_{k,i} \label{eq:mpc-f} \\
    &\forall k = 0, \ldots, N-1 \nonumber, 
    \forall i = 0, 1 \nonumber
    \end{align}
\end{subequations}

\noindent where $\decisionset$ is the full set of decision variables composed of the state $\state_k$ and control input $\inp_k$ at all timesteps $k = 0, \ldots, N$. 
The cost function includes the trajectory tracking error and input cost, where $\state_k^\text{ref} \in \R^3$ is the state reference, and $\Qweight \in \R^{13\times13}$ and $\Rweight \in \R^{12\times12}$ are diagonal positive semi-definite cost-weight matrices. 
The equality constraints include the discrete-time dynamics (\ref{eq:mpc-b}), where $\Amat$ and $\Bmat$ represent the discrete-time state-space matrices from Euler integration,
 and the contact constraint~(\ref{eq:mpc-c}), where $\contactstate$ and $\inp_{k, i}$ are the contact state ($0$ or $1$) and the contact wrench of the $i^{\rm th}$ foot at timestep $k$, respectively. 
The inequality constraints include the friction pyramid~(\ref{eq:mpc-d}), force saturation~(\ref{eq:mpc-e}), and line contact constraints~(\ref{eq:mpc-f}). 
Here, $\mu$ represents the friction coefficient; $\force_{k,i}$ and $\moment_{k,i}$ represent the ground reaction force and moment of the $i^{\rm th}$ foot at timestep $k$; $\mathrm{F}_{\max}$ represents the maximum vertical force; $l_t$ and $l_h$ denote the distances from the ankle joint to the toe and heel; $\SelMat_\mathrm{y} \in \R^{1\times3}$ and $\SelMat_\mathrm{z} \in \R^{1\times3}$ are row vectors that select the $y$ and $z$ elements, respectively; and $\FootRotMat$ is the rotation matrix of the $i^{\rm th}$ foot w.r.t the world frame.

\subsection{Low level control}
Low-level control follows existing literature~\cite{di2018dynamic,li2021force}, where the stance foot torque is computed from the contact Jacobian mapping, and the swing foot torque is calculated using a proportional–derivative (PD) controller whose target setpoints are generated via analytical inverse kinematics (IK). 
The analytical IK computes joint position commands corresponding to the desired foot position $\pfdes$, which is derived from a heuristic swing leg trajectory whose end point is specified by the foot placement planner. 
For the foot placement planner, we use Raibert heuristics~\cite{raibert1989dynamically} to update the target footholds $\foothold \in \R^3$ at each control timestep:

\begin{equation}
    \foothold = \phip^\text{ref} + \textstyle{\frac{1}{2}}\linvel\delta{T} + k_d(\linvel-\vdes),
\end{equation}
where $\mathbf{p}^\text{ref}_\text{hip} \in \R^3$ represents the hip position reference; $\delta{T}$ denotes the remaining step duration in a single walking step; $\mathbf{\dot{p}}^\text{ref} \in \R^3$ is the reference CoM velocity in the world frame; and $k_d$ is a feedback gain (set to $0.05$ in this work).
Finally, the reference swing foot trajectory $\pfdes \in \R^3$ is computed using a cubic Bézier curve $\Bezier$, where $\p_0$ is the stance foot position right before contact switch,  and $\p_1$ and $\p_2$ are the control points given by
$\p_1 = \p_0 + \tfrac{1}{3}(\foothold - \p_0), \quad \p_2 = \p_0 + \tfrac{2}{3}(\foothold - \p_0)$, and $\phi_t$ is a swing phase variable that is mapped to a predefined swing state ($0$ or $1$).
At each control step, $\phi_t$ is updated by $dt/T_\text{s}$, where $dt$ is the control timestep and $T_\text{s}$ is the swing phase duration.

\section{Methodology} 
In this section, we introduce our RL-augmented MPC, where the RL policy augments the system dynamics, swing leg trajectory, and step duration. 
The following subsections describe how the RL policy augments each component. 


\newcommand*{\highlight}{\textcolor[RGB]{0, 102, 102}}

\newcommand{\resdynamics}{\mathbf{f}_\text{r}(\mathbf{x}, \mathbf{u})}
\newcommand{\drift}{\mathbf{\hat{f}_x(x)}}
\newcommand{\affine}{\mathbf{\hat{f}_u(x)}}
\newcommand{\ResAcc}{\mathbf{\hat{f}^{(1)}_x}}
\newcommand{\ResAngAcc}{\mathbf{\hat{f}^{(2)}_x}}
\newcommand{\ResInvMass}{\mathbf{\hat{f}^{(1)}_u}}
\newcommand{\ResInvInertia}{\mathbf{\hat{f}^{(2)}_u}}
\newcommand{\SimpleResInvMass}{\mathrm{diag}(\hat{\mathrm{f}}^{(1)}_{\mathrm{u}11}, \hat{\mathrm{f}}^{(1)}_{\mathrm{u}22}, \hat{\mathrm{f}}^{(1)}_{\mathrm{u}33})}
\newcommand{\SimpleResInvInertia}{\mathrm{diag}(\hat{\mathrm{f}}^{(2)}_{\mathrm{u}11}, \hat{\mathrm{f}}^{(2)}_{\mathrm{u}22}, \hat{\mathrm{f}}^{(2)}_{\mathrm{u}33})}
\newcommand{\resdynamicsets}{[
\ResAcc, 
\ResAngAcc, 
\hat{\mathrm{f}}^{(1)}_{\mathrm{u}11}, 
\hat{\mathrm{f}}^{(1)}_{\mathrm{u}22}, 
\hat{\mathrm{f}}^{(1)}_{\mathrm{u}33}, 
\hat{\mathrm{f}}^{(2)}_{\mathrm{u}11}, 
\hat{\mathrm{f}}^{(2)}_{\mathrm{u}22}, 
\hat{\mathrm{f}}^{(2)}_{\mathrm{u}33}]}
\newcommand{\pz}{\mathrm{h}}
\newcommand{\deltapz}{\Delta{\mathrm{h}}}
\newcommand{\nominalpz}{\widetilde{\mathrm{h}}}
\newcommand{\deltacp}{\Delta_\text{cp}}
\newcommand{\dtmpc}{dt_\text{mpc}}
\newcommand{\dtmpcnominal}{\widetilde{dt}_{\text{mpc}}}

\subsection{System dynamics adaptation}
Single rigid-body dynamics (SRBD) represents a reduced-order model with several simplifications: the leg masses are ignored, and the robot’s centroidal inertia does not change drastically~\cite{di2018dynamic}.
In addition, the robot's center of mass experiences various disturbances, including foot–terrain interactions and inertial forces from the swing legs \cite{gu2025robust}. 
While MPC with whole-body dynamics offers a potential solution, this approach incurs significant computational costs and implementation complexity. 
Therefore, we maintain the use of simplified dynamics, but augment this model using RL. 
Our baseline MPC is solved at $100$ Hz; consequently, we retrieve residual dynamics from RL at the same frequency.

Following existing literature~\cite{pandala2022robust, chen2024learning-usc, kim2025modular}, we define residual dynamics $\resdynamics$ as offset terms to the linear and angular accelerations that account for unmodeled foot inertia and ground impact effects. 
We express these offset terms in an affine control system~\cite{brunke2022safe}, a standard model formulation, as
\begin{equation}
    \begin{array}{l}
        \dot{\mathbf{x}} = \dynamics + \resdynamics,  \mathbf{f}_\text{r}=
        \begin{bmatrix}
            \mathbf{0_{6\times1}}
            \\ 
            \drift + \affine\inp
        \end{bmatrix} 
    \end{array}
    \label{eq:augmented-srbd}
\end{equation}
where $\drift\in \R^6$ represents the residual state term, and $\affine\in \R^{6\times12}$ denotes the residual actuation matrix term. 
The first six rows of $\resdynamics$ are zeros, since the residual dynamics only model perturbations at the acceleration level. 

Furthermore, we simplify the $\affine$ matrix by assuming that moments do not affect linear acceleration and forces do not affect angular acceleration. 
This simplification reduces the original actuation matrix $\affine\in \R^{6\times12}$ to two low-dimensional matrices $\ResInvMass, \ResInvInertia\in\R^{3\times3}$.
Then the linearized augmented dynamics are further simplified as
\begin{equation}
    \begin{array}{l}
        \dot{\mathbf{x}} = \nominaldynamics + 
        \begin{bmatrix}
            \mathbf{0_{6\times1}}
            \\ 
            \highlight{\ResAcc} + \highlight{\ResInvMass} \sum_i \force_i
            \\
            \highlight{\ResAngAcc} + \highlight{\ResInvInertia} \sum_i \moment_i
        \end{bmatrix}
        \\
    \end{array}
    \label{eq:augmented-srbd}
\end{equation}
where $\nominaldynamics$ denotes the linearized single rigid body dynamics from the previous section, $\highlight{\ResAcc},\highlight{\ResAngAcc} \in \R^3$ are residual accelerations, and $\highlight{\ResInvMass}, \highlight{\ResInvInertia} \in  \R^{3\times3}$ are simplified residual actuation matrices.
We further simplify the matrices $\ResInvMass, \ResInvInertia$ as $\SimpleResInvMass \in \R^{3\times3}$ and $\SimpleResInvInertia \in \R^{3\times3}$ by ignoring off-diagonal elements to reduce the number of variables from 9 to 3 for each matrix. 
Consequently, the residual dynamics parameters to be learned are $\resdynamicsets \in \R^{12}$. 


\subsection{Swing leg trajectory adaptation}

Designing collision-free swing leg trajectories is critical, especially for locomotion on non-flat terrain. 
To this end, we design an adaptive swing leg trajectory by adjusting its apex height and control points, as illustrated in Fig.~\ref{fig:fig2}. 
Specifically, we modify the cubic Bézier curve $\Bezier$ using the adaptive control points $\p_1, \p_2$:

\begin{equation}
    \begin{aligned}
         \p_1 &= \p_0 + (\tfrac{1}{3}+\highlight{\deltacp})(\foothold - \p_0)
         \\
         \p_2 &= \p_0 + (\tfrac{2}{3}+\highlight{\deltacp})(\foothold - \p_0)
         \\
         \mathrm{p_1^z} &=  \mathrm{p_2^z} = \tfrac{1}{6}(8 \pz - \mathrm{p_0^z} -\mathrm{\mathrm{p^z_\text{foothold}}})
         \\
          \pz &=  \mathrm{p_0^z} + \nominalpz + \highlight{\deltapz}
    \end{aligned}
\end{equation}
where $\p_0$ is the stance foot position right before contact switch, $\foothold$ is the planned foothold, $\pz$ is the apex height of reference foot trajectory, $\nominalpz$ is the nominal swing foot height.  
We introduce two residual parameters: the residual apex height  $\highlight{\deltapz}$ and the control point coefficient $\highlight{\deltacp}$, which modify the shape of the swing leg trajectory.
These two parameters adjust the foot's reference height and the position of the apex, respectively, as illustrated in Fig.~\ref{fig:fig2}. 
A positive $\deltacp$ shifts the trajectory forward, whereas a negative value shifts it backward.
This formulation enables flexible manipulation of the trajectory while maintaining a small search space, compared to approaches that learn residuals for the entire joint PD setpoints~\cite{chen2024learning-usc}.

\subsection{Gait frequency adaptation}
Our controller uses a periodic gait to pre-generate the contact sequence for both legs. 
We achieve variable walking gaits by parameterizing the swing phase duration as $T_s = Hdt_\text{mpc}$, where $H$ represents MPC's horizon length and $\dtmpc$ denotes the discretization timestep (or sampling time).
Therefore, the step duration can be adjusted by manipulating the MPC sampling time.
To this end, we formulate $\dtmpc$ as follows:
\begin{equation}
    \dtmpc = \dtmpcnominal(1 + \highlight{s})
\end{equation}
where $\dtmpcnominal$ is the nominal MPC sampling time, and $\highlight{s}$ is the MPC sampling time coefficient. 
The double-support and single-support durations are computed as $T_{DS}=2\dtmpc$ and $T_{SS}=8\dtmpc$, respectively.
Thus, the residual double-support and single-support durations are $\Delta{T_{DS}}=2s\dtmpcnominal$ and $\Delta{T_{SS}}=8s\dtmpcnominal$, respectively.
The MPC then uses the updated double-support and single-support durations to generate the contact sequence at each RL step.
We assume these durations remain constant during the MPC prediction horizon, since the horizon is short ($0.2 \sim 0.3$ s) and the MPC solution is updated at a high frequency ($100$ Hz). 

\subsection{RL formulation}
\newcommand{\action}{\mathbf{a_t}}
\newcommand{\prevaction}{\mathbf{a_{t-1}}}
\newcommand{\obs}{\mathbf{o_t}}
\newcommand{\reward}{r_t}

The objective of the RL policy is to estimate the residual action parameters discussed above.
We treat the problem of finding these residual parameters as a Markov Decision Process (MDP). 
At each timestep $t$, the agent receives the observation $\mathbf{o}_t$, performs an action $\mathbf{a}_t$ according to its current policy $\pi(\cdot|\mathbf{o}_t)$, and obtains a scalar reward $R_t$. 
The goal of the agent is to maximize the discounted sum of rewards $\mathbb{E}_\pi [\sum_{t=0}^\infty \gamma^t R_t]$, where $\gamma\in [0,1)$ is the discount factor. 

We apply model-free deep reinforcement learning (DRL) methods to train a neural network that maximizes the return and finds the optimal residuals for the MPC.
The predicted residual parameters $\mathbf{a}_t\sim \pi(\cdot|\mathbf{o}_t)$ augment the MPC controller to enable adaptive locomotion.
We leverage the robotics simulation platform NVIDIA IsaacLab~\cite{mittal2023orbit} with its state-of-the-art physics engine for whole-body dynamics simulation.

\textbf{Action space:} 
The action space $\mathcal{A}$ consists of dynamics residuals $\resdynamicsets \in \R^{12}$, swing foot controller residuals $[\deltapz, \deltacp] \in \R^2$, and the MPC sampling time coefficient $s \in \R$.
The actions from the policy are scaled to match the range of each residual term using the scaling parameters listed in Table~\ref{tab:action}.

\begin{table}[t]
    \centering
    \caption{List of action space. 
    }
    \begin{minipage}{1.0\linewidth}
        \centering
        \begin{tabular}{ccc}
        \toprule
        \textbf{Action terms} & \textbf{Scaling parameters} & \textbf{Unit} \\
        \midrule
        Res. lin. acc. $\ResAcc$ & ($2.0, 2.0, 4.0$) & ($\mathrm{m/s^2}$)
        \\ 
        Res. ang. acc. $\ResAngAcc$ & ($1.0, 1.0, 1.0$) & $(\mathrm{rad/s^2})$
        \\
        Res. act. mat. $\ResInvMass$ & ($\tfrac{0.2}{13.856}, \tfrac{0.2}{13.856}, \tfrac{0.2}{13.856}$) & ($\mathrm{/kg}$)
        \\
        Res. act. mat. $\ResInvInertia$ & ($\tfrac{0.2}{0.5413}, \tfrac{0.2}{0.52}, \tfrac{0.2}{0.0691}$) & ($\mathrm{/kg\cdot m^2}$)
        \\
        Sampling time coef. $s$ & $0.3$ & -
        \\
        Swing foot $\deltapz$ & $0.05$ (slippery), $0.15$ (rough) & $(\mathrm{m})$
        \\
        Control point $\deltacp$ & $0.33$ (slippery), $0.66$ (rough) & -
        \\
        \bottomrule
        \end{tabular}
        \vspace{-3mm}
    \end{minipage}
    \label{tab:action}
\end{table}

\textbf{Observation space:}
The observation space consists of proprioceptive sensory data from the robot and foot states from the MPC controller. 
Proprioceptive information includes base linear velocity $\linvel$ and angular velocity $\angvel$, joint positions $\mathbf{p}_{\text{joint}}$, and joint velocities $\dot{\mathbf{p}}_{\text{joint}}$. 
Additionally, we include the command velocity and the computed projected gravity $\hat{\mathbf{g}}$ based on IMU data. 
Foot states consist of the current foot positions $\pfoot$, reference foot positions $\pfdes$, planned footholds $\foothold$, and the swing phase $\phi$. 
Finally, the actions from the previous timestep $\mathbf{a}_{t-1}$ are also incorporated into the observation, which has been shown to produce smooth actions~\cite{rudin2022learning}. 

\begin{table}[t]
    \centering
    \scriptsize
    \caption{List of observation space.}
    \begin{minipage}{0.9\linewidth}
        \centering
        \begin{tabular}{cc|cc}
        \hline
        \textbf{Observation} & \textbf{Dim} & \textbf{Observation} & \textbf{Dim} \\
        \hline
        Projected gravity & 3 & Base linear vel. & 3 \\
        Base angular vel. & 3 & Velocity commands & 3 \\
        Joint pos. & 10 & Joint vel. & 10 \\
        Previous action & 9 & Swing phase & 2 \\
        Planned footholds & 4 & Foot pos. & 6 \\
        Reference foot pos. & 6 \\
        \hline
        \end{tabular}
        \vspace{-3mm}
    \end{minipage}
    \label{tab:observation}
\end{table}


\newcommand{\vz}{\dot{\mathrm{p}}_\mathrm{z}}
\newcommand{\wxy}{{\omega_\mathrm{x}}^2 + {\omega_\mathrm{y}}^2}
\newcommand{\tanfeetvel}{\|\dot{\mathbf{p}}_{\text{foot}, i}\|^2}
\newcommand{\leftfootypos}{p_{\text{f},0,y}}
\newcommand{\rightfootypos}{p_{\text{f},1,y}}
\newcommand{\stepwidth}{d}

\textbf{Reward functions:}
Our reward design is summarized in Table~\ref{tab:reward}.
We leverage existing reward functions implemented in IsaacLab for bipedal locomotion tasks, which include velocity command tracking, center of mass height tracking, and others.
Reward functions specific to our problem include penalties on the base-to-stance-foot angle $\theta_\text{foot}$ and the step width $\stepwidth$.

\begin{table}[t]
    \centering
    \scriptsize
    \caption{List of reward functions.}
    \begin{minipage}{0.9\linewidth}
        \centering
        \begin{tabular}{ccc}
        \toprule
        \textbf{Reward} & \textbf{Expression} & \textbf{Weight} \\
        \midrule
        Track lin. vel. XY & $\exp\left(-\|\linvel - \vdes\|^2 / {\sigma^2}\right)$ & 1.0 \\
        Track ang. vel. Z & $\exp\left(-\|\angvel - \angvel^\text{ref}\|^2 / {\sigma^2}\right)$ & 0.5 \\
        Track height & $\exp\left(-(\mathrm{p^z} - \zdes)^2 / {\sigma^2}\right)$ & 0.1 \\
        Lin. vel. Z & $\vz$ & -1e-2 \\
        Ang. vel. XY & $\wxy$ & -1e-4 \\
        Joint vel. & $\|\mathbf{\dot{q}}_\text{joint}\|^2$ & -2.5e-4 \\
        Action smoothness & $\| \action - \prevaction \|^2$ & -0.015 \\
        Feet slide & $\sum_i \mathrm{c}_i \tanfeetvel$ & --0.01 \\
        Knee collision & ${1}_\text{knee}$ & -5.0 \\
        Leg-base angle & $\theta_\text{leg}^2$ & -1.0 \\
        Step width & $\stepwidth^2$ & -0.2 \\
        \bottomrule
        \end{tabular}
    \end{minipage}
    \vspace{-2mm}
    \label{tab:reward}
\end{table}

\begin{table}[t]
    \caption{List of MPC parameters.}
    \begin{minipage}{0.9\linewidth}
        \centering
        \scriptsize
        \begin{tabular}{c@{\hspace{1pt}}c}
        \toprule
        \textbf{Name} & \textbf{Value} \\
        \midrule
        Horizon length $H$ & $10$ \\
        Sampling time $\dtmpcnominal$ & $0.025$ s \\
        Swing height $\widetilde{\mathrm{h}}$ & $0.1$ m \\
        Toe length $l_t$ & $0.07$ m \\
        Heel length $l_h$ & $0.04$ m \\
        Max. force $F_{\max}$ & $500$ N \\
        State weight $\mathbf{Q}$ & $\text{diag}(150,150,250,100,100,250,1,1,1,10,10,1,1)$ \\
        Control weight $\mathbf{R}$ & \text{diag}($10^{-5}, \ldots, 10^{-4}, \ldots, 10^{-4}$) \\
        \bottomrule
        \end{tabular}
        \vspace{-5mm}
    \end{minipage}
    \label{tab:mpc-params}
\end{table}

\textbf{Implementation detail:}
In this work, we employ Soft Actor-Critic~\cite{haarnoja2018soft}, an off-policy model-free RL algorithm.
An off-policy algorithm is suitable for our approach because it is more sample efficient~\cite{haarnoja2018soft}, which is crucial since environment rollouts with MPC are computationally expensive.
Both the actor and critic networks are implemented as separate three-layer MLPs with hidden sizes of ($512$, $256$, $128$) with the ELU activation function.
The policy is parameterized as a Gaussian distribution during training, with the mean and standard deviation predicted by a single MLP.

\textbf{Training:}
The RL policy is trained on an NVIDIA RTX 4090 using IsaacLab~\cite{mittal2023orbit}.
Training is performed using the SAC algorithm implemented in \texttt{rl\_games}~\cite{rl-games2021}, a GPU-accelerated RL library.
The hyperparameters used during training are provided in the open-source code.
During both training and inference, MPC and RL run at $100$ Hz, while low-level control and physics simulation run at $400$ Hz.
The parameters of MPC are listed in Table~\ref{tab:mpc-params}.

Given this setup, the policy often converged to suboptimal behavior, outputting saturated actions most of the time. 
To mitigate this poor exploration, we introduce a squared action loss implemented in \texttt{rl\_games}. 
With this loss, our policy does not saturate. 



\newcommand{\srmpc}{\text{SR}_\text{MPC}\!\uparrow \mathrm{(\%)}}
\newcommand{\srours}{\text{SR}_\text{ours}\!\uparrow \mathrm{(\%)}}
\newcommand{\evmpc}{e^{v}_\text{MPC}\!\downarrow \mathrm{(m/s)}}
\newcommand{\evours}{e^{v}_\text{ours}\!\downarrow \mathrm{(m/s)}}
\newcommand{\erollmpc}{e^{\theta_x}_\text{MPC}\!\downarrow \mathrm{(deg)}}
\newcommand{\erollours}{e^{\theta_x}_\text{ours}\!\downarrow \mathrm{(deg)}}
\newcommand{\epitchmpc}{e^{\theta_y}_\text{MPC}\!\downarrow \mathrm{(deg)}}
\newcommand{\epitchours}{e^{\theta_y}_\text{ours}\!\downarrow \mathrm{(deg)}}

\begin{table*}
    \centering
    \caption{Comparison between the baseline MPC and our approach. 
    The table shows success rate (SR), forward velocity tracking error ($e^v$), and roll/pitch errors ($e^{\theta_x}, e^{\theta_y}$) collected from $100$ episodes on rough terrain and slippery terrain with varying levels of difficulty. 
    The highest success rate and the lowest tracking errors are highlighted in bold.
    }
    \begin{minipage}{1.0\linewidth}
        \centering
        \scriptsize
        \begin{tabular}{c@{\hspace{3pt}}c||c@{\hspace{3pt}}c||c@{\hspace{3pt}}c||c@{\hspace{3pt}}c||c@{\hspace{3pt}}c}
        \toprule
        Env & height ($\mathrm{cm}$) / $\mu$ (-) & $\srmpc$ & $\srours$ & $\evmpc$ & $\evours$ & $\erollmpc$ & $\erollours$ & $\epitchmpc$ & $\epitchours$
        \\
        \midrule
        
        \multirow{3}{*}{\makecell{pyramid \\ stairs \\ (height)}} & $8$ & $100 \pm 0$ & $100 \pm 0$ & $0.12 \pm 0.09$ & $0.12 \pm 0.09$ & $\boldsymbol{0.55 \pm 0.38}$ & $0.55 \pm 0.40$ & $1.71 \pm 0.88$ & $\boldsymbol{1.6 \pm 0.94}$
        \\ 
        & $9$ & $48 \pm 5$ & $\boldsymbol{96 \pm 1.96}$ & $0.15 \pm 0.12$ & $\boldsymbol{0.14 \pm 0.1}$ & $0.68 \pm 1.03$ & $\boldsymbol{0.59 \pm 0.46}$ & $1.96 \pm 1.58$ & $\boldsymbol{1.68 \pm 1.1}$
        \\ 
        & $10$ & $1 \pm 0.99$ & $\boldsymbol{86 \pm 3.47}$ & $0.22 \pm 0.17$ & $\boldsymbol{0.16 \pm 0.13}$ & $0.97 \pm 1.87$ & $\boldsymbol{0.63 \pm 0.58}$ & $2.6 \pm 3.03$ & $\boldsymbol{1.7 \pm 1.16}$
        \\
        \midrule
        
        \multirow{3}{*}{\makecell{random \\ stairs \\ (height)}} & $8$ & $99 \pm 0.99$ & $99 \pm 0.99$ & $0.1 \pm 0.09$ & $0.1 \pm 0.09$ & $0.57 \pm 0.39$ & $\boldsymbol{0.56 \pm 0.4}$ & $1.39 \pm 0.84$ & $\boldsymbol{1.32 \pm 0.88}$
        \\ 
        & $9$ & $69 \pm 4.62$ & $\boldsymbol{98 \pm 1.4}$ & $0.12 \pm 0.11$ & $\boldsymbol{0.12 \pm 0.1}$ & $0.66 \pm 0.89$ & $\boldsymbol{0.58 \pm 0.43}$ & $1.59 \pm 1.51$ & $\boldsymbol{1.37 \pm 0.95}$
        \\ 
        & $10$ & $15 \pm 3.57$ & $\boldsymbol{85 \pm 3.57}$ & $0.19 \pm 0.15$ & $\boldsymbol{0.12 \pm 0.11}$ & $0.96 \pm 1.94$ & $\boldsymbol{0.61 \pm 0.48}$ & $2.08 \pm 2.58$ & $\boldsymbol{1.47 \pm 1.18}$
        \\
        \midrule
        
        \multirow{3}{*}{\makecell{stepping \\ stones \\ (height)}} & $\mathcal{U}(-5, 5)$ & $100 \pm 0$ & $100 \pm 0$ & $0.09 \pm 0.08$ & $0.09 \pm 0.08$ & $0.56 \pm 0.39$ & $\boldsymbol{0.54 \pm 0.34}$ & $1.24 \pm 0.64$ & $\boldsymbol{1.18 \pm 0.6}$
        \\ 
        & $\mathcal{U}(-6, 6)$ & $86 \pm 3.47$ & $\boldsymbol{96 \pm 1.96}$ & $0.1 \pm 0.09$ & $0.1 \pm 0.09$ & $0.6 \pm 0.59$ & $\boldsymbol{0.57 \pm 0.49}$ & $1.35 \pm 1.14$ & $\boldsymbol{1.25 \pm 0.87}$
        \\ 
        & $\mathcal{U}(-7, 7)$ & $68 \pm 4.66$ & $\boldsymbol{89 \pm 3.13}$ & $0.12 \pm 0.11$ & $\boldsymbol{0.11 \pm 0.09}$ & $0.67 \pm 1.0$ & $\boldsymbol{0.61 \pm 0.68}$ & $1.58 \pm 1.82$ & $\boldsymbol{1.28 \pm 0.95}$
        \\
        \midrule

        \multirow{3}{*}{\makecell{slippery \\ surface \\ ($\mu$)}} & $0.2$ & $100 \pm 0$ & $100 \pm 0$ & $0.09 \pm 0.08$ & $\boldsymbol{0.09 \pm 0.07}$ & $\boldsymbol{0.59 \pm 0.29}$ & $0.73 \pm 0.34$ & $\boldsymbol{1.11 \pm 0.43}$ & $1.4 \pm 0.38$
        \\
        & $0.15$ & $81 \pm 3.92$ & $\boldsymbol{100 \pm 0}$ & $0.1 \pm 0.08$ & $\boldsymbol{0.1 \pm 0.07}$ & $0.79 \pm 0.94$ & $\boldsymbol{0.65 \pm 0.34}$ & $\boldsymbol{1.26 \pm 0.76}$ & $1.41 \pm 0.4$
        \\
        & $0.1$ & $19 \pm 3.92$ & $\boldsymbol{100 \pm 0}$ & $0.16 \pm 0.12$ & $\boldsymbol{0.11 \pm 0.07}$ & $1.56 \pm 2.97$ & $\boldsymbol{0.61 \pm 0.37}$ & $1.55 \pm 1.7$ & $\boldsymbol{1.42 \pm 0.42}$
        \\
        & $0.05$ & $5 \pm 2.18$ & $\boldsymbol{74 \pm 4.39}$ & $0.22 \pm 0.16$ & $\boldsymbol{0.13 \pm 0.09}$ & $1.95 \pm 3.6$ & $\boldsymbol{0.78 \pm 1.29}$ & $1.56 \pm 1.84$ & $\boldsymbol{1.48 \pm 0.71}$
        \\
        \bottomrule
        \end{tabular}
    \end{minipage}
    \label{tab:benchmark}
    \vspace{-3mm}
\end{table*}

\section{Evaluation}

\subsection{Terrain setups}
\begin{enumerate}
    \item Pyramid stairs: 
    This type of terrain contains a set of ascending steps arranged in all four directions. 
    Each step has a width of $25$ cm and a height of $10$ cm. 

    \item Random stairs: 
    This type of terrain contains a set of steps with a random ascending/descending pattern. 
    The width and height of each step are the same as the pyramid stairs. 

    \item Stepping stones:
    This terrain is constructed by extruding or intruding a flat grid cell by a random height.
    Each grid has a width of $25$ cm, and the height is uniformly sampled from $\mathcal{U}(-h_{\max}, h_{\max})$ with $h_{\max}=7$ cm.

    \item Slippery surface: 
    This type of terrain contains a set of row-indexed sub-terrains, where the friction range decreases monotonically with the row index.
    In each sub-terrain, the friction coefficient is randomly assigned either a high value ($\mu = 0.5$) or a low value sampled from the predefined range.
    As the row index increases, the minimum friction value decreases to 0.05.
\end{enumerate}
During training, a terrain curriculum is used, adapting the robot’s initial position to a more difficult region, depending on whether walking succeeds at the current region. 
For example, on slippery terrain, if the robot can walk in the current region, it is moved to a lower-friction region.

\subsection{Evaluation metrics}
We use success rate (SR), velocity/orientation tracking accuracy $e^v, e^{\theta_x}, e^{\theta_y}$ as metrics with a thorough analysis of the action outputs.
The success rate is measured by running $100$ episodes and counting the number of trials that succeed in reaching the end of a terrain.
An episode is counted a failure if the robot violates height constraints ($p_z \geq 0.4$), orientation constraints ($\theta_x,\theta_y \in (-\tfrac{\pi}{6}, \tfrac{\pi}{6})$), or fails to reach the end of the terrain within $20$ s.
Tracking accuracy is measured as the absolute error of the base velocity, which evaluates the policy’s command tracking capability, with episodes run for $6$ s.
Finally, we analyze the action outputs to assess the contribution of the RL policy.
A successful policy should output near-zero actions under nominal conditions, such as on flat terrain with a normal frictional coefficient, and produce large corrective actions on slippery surfaces and rough terrain.

\section{Results}

\newcommand{\sr}{\text{SR}\!\uparrow \mathrm{(\%)}}
\newcommand{\ev}{{e^v}\!\downarrow \mathrm{(m/s)}}
\newcommand{\eroll}{e^{\theta_x}\!\downarrow \mathrm{(deg)}}
\newcommand{\epitch}{e^{\theta_y}\!\downarrow \mathrm{(deg)}}

In this section, we evaluate the performance of the proposed algorithm for robust walking on different types of terrain. 
The robot's state is initialized with the values listed in Table~\ref{tab:initialization} at the beginning of each episode. 

\begin{table}[h]
    \caption{Initialization parameters. $U(a)$ denotes the uniform distribution $\mathcal{U}(-a, a)$. 
    }
    \scriptsize
    \begin{minipage}{1.0\linewidth}
        \centering
        \begin{tabular}{cccc}
        \toprule
        {} & \textbf{Name} & \textbf{Value} & \textbf{Unit} \\
        \midrule
        Training & Pos. $\p$ &($U(2.0), U(2.0), 0.55$) & ($\mathrm{m}$) \\
        & Ori. $\ori$ & ($0,  0, U(\pi)$) & ($\mathrm{rad}$) \\
        & Cmd. $\mathrm{\dot{p}}^{\text{ref},\mathrm{x}}$ & $0.5$ & ($\mathrm{m/s}$) \\
        \hline
        Inference & Pos. $\p$ & ($U(0.3), U(0.3), 0.55$) & ($\mathrm{m}$) \\
        pyramid stairs & Ori. $\ori$ & ($0, 0, \pm \tfrac{n\pi}{2}$) & ($\mathrm{rad}$) \\
        {random stairs} & {Ori. $\ori$} & ($0, 0, \pm n\pi$) & ($\mathrm{rad}$) \\
        {stepping stones} & {Ori. $\ori$} & ($0, 0, U(\pi)$) & ($\mathrm{rad}$) \\
        {slippery surface} & {Ori. $\ori$} & ($0, 0, U(\pi)$) & ($\mathrm{rad}$) \\
        {} & Cmd. $\mathrm{\dot{p}}^{\text{ref},\mathrm{x}}$ & $0.5$ & ($\mathrm{m/s}$) \\
        \bottomrule
        \end{tabular}
    \end{minipage}
    \vspace{-5mm}
    \label{tab:initialization}
\end{table}

\begin{figure*}[t]
    \centering
    \begin{minipage}{0.47\linewidth}
        \hspace{-6mm}
        \includegraphics[width=1.0\linewidth]{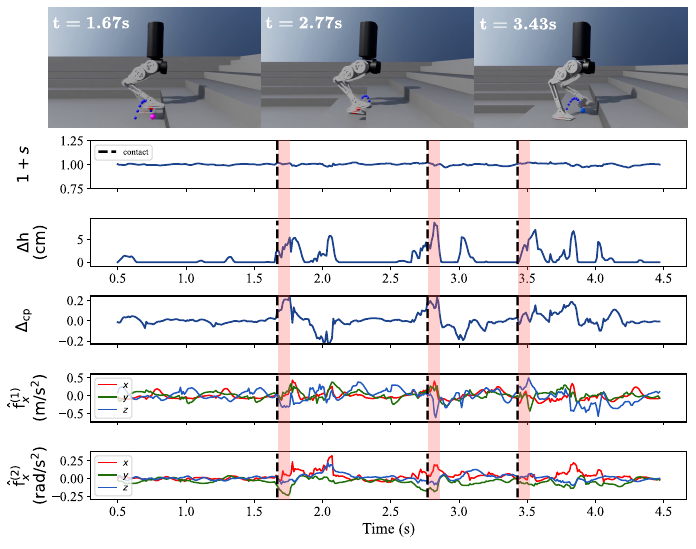}
        \label{fig:rl-action-inv-pyramid}
        \vspace{-3mm}
    \end{minipage}
    \begin{minipage}{0.47\linewidth}
        \hspace{-2mm}
        \includegraphics[width=1.0\linewidth]{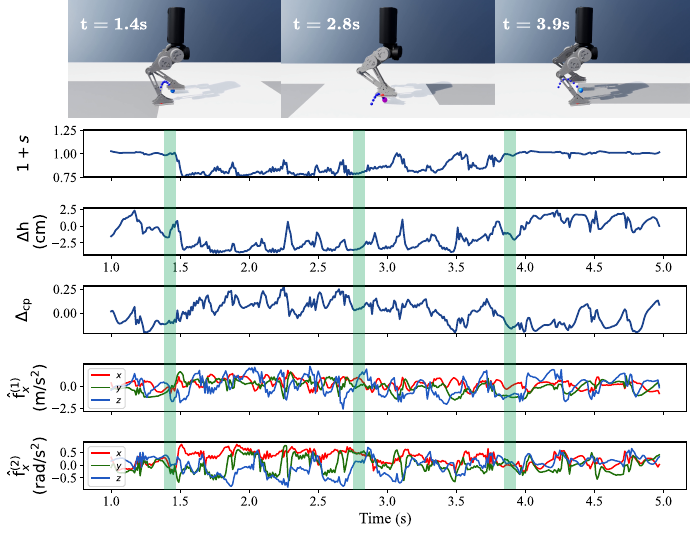}
        \label{fig:rl-action-slip-terrain}
        \vspace{-3mm}
    \end{minipage}
    \caption{
    The figure illustrates actions taken by the RL policy and time-lapse images of the robot at times marked by red (rough terrain instances) and green (slippery terrain instances) shaded regions. 
    The action plots illustrate the outputs of each term on the pyramid stairs (left) and slippery surface (right). 
    Dotted vertical lines in the left figure indicates the instance of toe contact. 
    }
    \label{fig:rl-contribution}
    \vspace{-3mm}
\end{figure*}

\subsection{Per-terrain evaluation}
We evaluate performance across four terrain types, varying step height for rough terrains and minimum friction for slippery terrain.  
Table~\ref{tab:benchmark} summarizes benchmark results under various terrain types.
Across all terrain types, our method achieves substantially higher success rates and higher tracking accuracy compared to the MPC baseline. 

\noindent \textbf{Pyramid stairs: }
As the step height increases, the performance gap between the baseline controller and our approach becomes more pronounced. 
The baseline controller rapidly deteriorates, achieving only a $ 1\%$ success rate on $10$ cm stairs, indicating its limited robustness to challenging terrain. 
In contrast, our method sustains a success rate exceeding $ 85\%$, highlighting its ability to reliably handle significant terrain variations.
In addition to robustness, our method maintains better tracking performance across different levels of terrain difficulty. 
The variation in tracking error remains within $0.04$ m/s between mild and harsh terrain, while the baseline shows much larger deviations, up to $0.1$ m/s, further underscoring the advantage of our method.

\noindent \textbf{Random stairs: }
Overall, the success rate is slightly lower than on pyramid stairs due to the random descending–ascending transitions, where overspeed during descent increases the risk of tripping on the subsequent step. 
Even at the highest difficulty, our method sustains a success rate above $80\%$, whereas the baseline falls below $20\%$.  
Tracking error remains small across stair heights for our approach, in contrast to the baseline.

\noindent \textbf{Stepping stones: }
Our method consistently outperforms the baseline in success rate. 
Tracking errors are comparable between the two methods across all stone spacings, reflecting the relatively lower difficulty of stepping stones.  

\noindent \textbf{Slippery surface: }  
As the friction coefficient $\mu$ decreases, the baseline drops to a $5$\% at $\mu=0.05$, whereas our method maintains above $70\%$. 
Tracking accuracy shows a similar trend: our method limits error to $0.13$ m/s on the most slippery surface, while the baseline exceeds $0.2$ m/s.  
Moreover, our policy stabilizes torso roll angle more effectively, which is critical for resisting lateral slip.

\subsection{Role of RL Residuals}
We analyze the policy's corrective actions to illustrate how RL complements MPC on pyramid stairs and slippery surfaces, as presented in Fig.~\ref{fig:rl-contribution}.
Since the policy does not include perception in its observation, it reacts to the terrain only after the foot makes contact or experiences slippage.

On stairs, pronounced spikes appear in $\deltapz$, $\deltacp$, and $s$ at foot contact instants, indicated by the dotted vertical lines.  
When stepping onto stairs, $\deltapz$ increases by up to $10$ cm, raising the swing foot reference height to reduce the risk of tripping.  
Simultaneously, $\deltacp$ decreases to approximately $-0.2$, shifting the swing trajectory backward.  
This adaptation is critical for versatile and failure-free locomotion, as a symmetric trajectory ($\deltacp = 0$) often causes foot collisions when the prior foothold is near a stair edge. 
Residuals in linear and angular accelerations also increase, particularly in the vertical direction, thereby increasing the normal force at contact.  
In contrast, during steady walking on flat terrain ($t = 0.6$–$1.1$ s), residuals remain small and periodic, underscoring their adaptive, event-driven activation.  

On slippery terrain, both the amplitude and frequency of the residuals increase in response to reduced friction.  
At $t = 1.4$ s, the ratio of MPC sampling time to its nominal value decreases from $1.0$ to $0.75$, leading to shorter step durations.  
These shorter steps reduce stride length and mitigate lateral slip.  
Meanwhile, $\deltapz$ decreases to about $-2.5$ cm, lowering the swing foot to enable faster step-down motion.  
Linear and angular acceleration residuals adapt strongly, with pitch and yaw accelerations exhibiting the largest oscillations.

\subsection{Ablation study}
Our proposed method integrates multiple adaptation modules into the nominal MPC controller. 
To access the contribution of each module, we conduct an ablation study against eight variants, each representing a different combination of the adaptation modules. 
Specifically, \textit{res-all} corresponds to our proposed method with all three modules, \textit{vanilla-MPC} to the baseline MPC, and \textit{vanilla-RL} to a joint space RL policy trained with rewards implemented in IsaacLab for bipedal locomotion tasks.
Following the main evaluation metrics, we use success rate and velocity/orientation tracking as metrics, measured by running $100$ episodes on $10$ cm pyramid stairs. 
As summarized in Table~\ref{tab:ablation}, our method consistently outperforms the baseline MPC and RL controllers in terms of success rate and tracking accuracy.

The ablation study reveals specific weakness in the baselines. 
For example, the configurations without swing foot trajectory adaptation (\textit{dyn-gait}, \textit{dyn}, \textit{gait}) exhibit catastrophic failure, underscoring the critical role of swing leg adaptation on rough terrain. 
Furthermore, \textit{swing-gait} and \textit{dyn-swing} achieve lower success rates and larger tracking errors than \textit{swing} alone, highlighting that all three modules are essential for achieving the best performance. 

\begin{table}[h]
    \centering
    \scriptsize
    \caption{Results of the ablation study on $10$ cm pyramid stairs. The highest success rate and lowest tracking errors are highlighted in bold.
    }
    \begin{minipage}{1.0\linewidth}
        \centering
        \scriptsize
        \begin{tabular}{c@{\hspace{6pt}}c@{\hspace{3pt}}c@{\hspace{3pt}}c@{\hspace{3pt}}c}
        \toprule
        Name &  $\sr$ & $\ev$ & $\eroll$ & $\epitch$ \\
        \midrule
        res-all & $\boldsymbol{86 \pm 3.47}$ & $0.16 \pm 0.13$ & $\boldsymbol{0.63 \pm 0.58}$ & $\boldsymbol{1.7 \pm 1.16}$ \\
        swing & $74 \pm 1.0$ &  $0.17 \pm 0.14$ & $0.67 \pm 0.54$ & $1.72 \pm 1.13$ \\
        swing-gait & $67 \pm 4.7$ & $\boldsymbol{0.15 \pm 0.12}$ & $0.69 \pm 0.6$ & $1.91 \pm 1.36$ \\
        dyn-swing & $51 \pm 5.0$ & $0.18 \pm 0.15$ & $0.69 \pm 0.66$ & $1.85 \pm 1.48$ \\ 
        dyn-gait & $2 \pm 1.4$ & $0.21 \pm 0.17$ & $0.98 \pm 2.08$ & $2.53 \pm 2.82$ \\ 
        dyn & $1 \pm 0.99$ & $0.23 \pm 0.18$ & $1.02 \pm 2.08$ & $2.69 \pm 3.16$ \\ 
        gait & $1 \pm 0.99$ & $0.21 \pm 0.17$ & $0.9 \pm 1.64$ & $2.65 \pm 3.01$ \\
        vanilla-MPC & $1 \pm 0.99$ & $0.22 \pm 0.17$ & $0.97 \pm 1.87$ & $2.60 \pm 3.03$ \\
        vanilla-RL & $15 \pm 3.57$ & $0.18 \pm 0.15$ & $3.27 \pm 2.00$ & $5.85 \pm 2.29$ \\
        \bottomrule
        \end{tabular}
    \end{minipage}
    \vspace{-2mm}
    \label{tab:ablation}
\end{table}

\subsection{Batched MPC on GPU}
\begin{figure}[t]
    \centering
    \begin{minipage}{0.9\linewidth}
        \centering
        \includegraphics[width=1.0\linewidth]{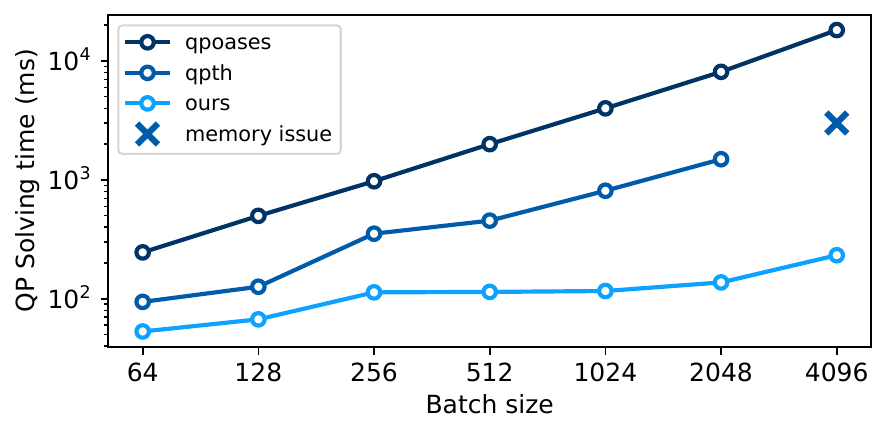}
        \vspace{-6mm}
    \end{minipage}
    \caption{Computational performance of \texttt{qpOASES}, \texttt{qpth}, and our method across different batch sizes in log scale. A cross mark denotes failure due to memory exhaustion when using \texttt{qpth}.
    }
    \label{fig:gpu-mpc-time-bench}
    \vspace{-5mm}
\end{figure}

Lastly, we present preliminary results of a batched MPC controller scalable to thousands of parallel environments, accelerating policy training by over $20$ times compared to a CPU baseline. 
We implement a sparse quadratic programming (QP) solver based on the primal-dual interior point method in CusADi~\cite{jeon2024cusadi} to enable training with a batch size of $4096$.
Fig.~\ref{fig:gpu-mpc-time-bench} presents benchmark results on computation time across different batch sizes. 
We compare our solver against \texttt{qpOASES}~\cite{qpoases}, a CPU-based QP solver used in the baseline MPC, and \texttt{qpth}~\cite{qpth}, a state-of-the-art batched QP solver implemented in PyTorch. 
As the batch size increases, our solver maintains significantly lower computation times, while both \texttt{qpOASES} and \texttt{qpth} exhibit substantial increases.
Furthermore, it achieves locomotion performance comparable to the CPU baseline, with the policy attaining a $85\%$ success rate on $10$ cm pyramid stairs. 
Given these performance gains, we are migrating our case studies to this GPU-accelerated implementation and will report updated results in future work.

\section{Conclusion}
In this work, we presented an RL-augmented MPC for robust bipedal locomotion over diverse terrains. 
Extensive simulation results demonstrate that our method achieves significantly higher success rates and more accurate tracking compared to the baseline.
Future research involves extensive hardware experiments and extending the current approach to perceptive locomotion. 




\bibliographystyle{IEEEtran}
\bibliography{reference.bib}

\end{document}